\documentclass[lettersize,journal]{IEEEtran}
\usepackage{amsmath,amsfonts, amssymb}
\usepackage{algorithmicx}
\usepackage{algorithm}
\usepackage{algpseudocode}
\usepackage{array}
\usepackage{titlesec}
\titlespacing*{\subsection}{0pt}{1em}{0.5em}

\usepackage{textcomp}
\usepackage{stfloats}
\usepackage{url}
\usepackage{verbatim}
\usepackage{graphicx}
\usepackage{subcaption}
\usepackage{float}
\usepackage{booktabs}
\usepackage{cite}
\usepackage{multirow}
\usepackage[margin=1in]{geometry}  
\begin{document}

\title{Few-Shot Concept Unlearning with Low Rank Adaptation}

\author{Shreyas Udaya, Aadarsh Lakshmi Narasiman}

\maketitle

\begin{abstract}



Image Generation models are a trending topic nowadays, with many people utilizing Artificial Intelligence models in order to generate images. There are many such models which, given a prompt of a text, will generate an image which depicts said prompt. There are many image generation models, such as Latent Diffusion Models, Denoising Diffusion Probabilistic Models, Generative Adversarial Networks and many more. 
When generating images, these models can generate sensitive image data, which can be threatening to privacy or may violate copyright laws of private entities. Machine unlearning aims at removing the
influence of specific data subsets from the trained models and in the case of image generation models, remove the influence of a concept such that the model is unable to generate said images of the concept when prompted. Conventional retraining of the model can take upto days, hence fast algorithms are the need of the hour. 
In this paper we propose an algorithm that aims to remove the influence of concepts in diffusion models through updating the gradients of the final layers of the text encoders. Using a weighted loss function, we utilize backpropagation in order to update the weights of the final layers of the Text Encoder componet of the Stable Diffusion Model, removing influence of the concept from the text-image embedding space, such that when prompted, the result is an image not containing the concept. The weighted loss function makes use of Textual Inversion and Low-Rank Adaptation.
We perform our experiments on Latent Diffusion Models, namely the Stable Diffusion v2 model, with an average concept unlearning runtime of 50 seconds using 4-5 images.

\end{abstract}


\section{Introduction} Machine unlearning is an emerging field that addresses the critical challenge of erasing the influence of specific data, concepts, or classes from a pre-trained machine learning model upon user request. While the database can be purged of personal data, the information contained therein may already have been assimilated into a model's parameters during training. This residual influence poses significant risks for privacy breaches and compliance violations under data protection regulations such as the General Data Protection Regulation (GDPR) \cite{voigt2017eu} and the California Consumer Privacy Act (CCPA) \cite{goldman2020introduction}. These regulations guarantee users the right to have their data removed, thus compelling organizations to develop effective machine unlearning techniques.


The advent of text-to-image models
generating high-quality images with the help of text prompts with some popular examples of these include - Stable Diffusion (SD) , Denoising Diffusion Probabilistic Model (DDPM) and DALL-E. These models are used in numerous applications \cite{lexica2025}\cite{novelai2025}\cite{picsart2025}\cite{firefly2025} related to art and design systems. With the popularity of these models there is a growing concern with regards to security, copyright issues, regulations, fairness etc. There is also concern about these models that generate biased and unsafe content.

Most of the data used in these text-to-image models while training includes a public dataset of web-scraped images and captions, which lack a guarantee on safety and bias, and private data for which there is no guarantee on quality.Therefore, data quality based filtering is unfeasible to address the unsafe and biased content, and at the same time, address concerns relating to privacy and copyright issues. Moreover, these models take days to train. 

Therefore, there is a need for methods that guide these large text-to-image models to forget concepts, which is called concept unlearning. One use case we address in this paper, for example, is unlearning the concept of Mickey Mouse, which may be needed in the case of avoiding copyright protection laws. 

Conventional retraining of these models is not feasible, however, due to the laborious amounts of time and computational resources it takes to train such models. 

There are many attempts at unlearning diffusion models. ESD \cite{gandikota2023erasing}, for example, proposes a fine-tuning method Erasing Stable Diffusion which modifies the weights of the U-Net model in order to reduce the probability of a concept of an image being generated. The authors further improve in UCE by updating the weights of the cross-attention in U-Net by closed-form. Unlike
the former method, it does not require the output of the Original Stable Diffusion Model
\cite{Fuchi_2024_BMVC} has developed a methodology targeting the text encoder for unlearning, where the authors use the concept of Textual Inversion \cite{gal2022image}, which is a training technique for fine-tuning image generation models by updating the text embeddings using a few images. The authors make use of negative loss with backpropagation in order to update the weights of the Text encoder model. However, this algorithm is still susceptible to generating images of the forget concept. 

Using this concept as a backbone, we propose a weighted loss function that makes use of few-shot unlearning along with Low-Rank Adaptation in order to perturb the image space and forget by using a projecting loss of forget embeddings onto the weights using backpropagation. This ensures better forgetting of the image space. However, the rank should not be so high as to forget the retain concepts, and not too low to not forget effectively. The optimal value depends on the influence of the forget concept.

The key contributions of this work are as follows: 

\begin{itemize} 
\item Propose an unlearning algorithm for Concept unlearning of the stable diffusion framework, such that it is unable to generate images of the forgotten concept.

\item Target the final encoder layers of the CLIP model, leaving less need to disturb the entire model.

\item Incorporate the entire diffusion model into the unlearning process and introduce Low rank Perturbation which we project onto the image space.

\end{itemize}
This paper is organized as follows. Section II reviews related work on machine unlearning and unlearning in Image Generation Models. Section III details our proposed methodology. Section IV presents the experimental results, and Section V concludes the paper with discussions on future research directions.

\section{Related Work}

Image Generation models are a subset of generative models that are capable of generating realistic images from inputs such as random noise and text prompts.These models can learn patterns, textures, and structures from the training dataset and use this knowledge to generate new images that resemble the training data or follow specific instructions. Some examples of the popular Image Generative models include - Generative Adversial Networks (GANs), Variational Autoencoders (VAEs) and Diffusion Models.
Of these various types of Generative Models, Diffusion Models create high quality data by reversing the gradual process of adding noise. Some of the popular Diffusion Models include - Denoising Diffusion Probabilistic Models (DDPM), Stable Diffusion Models, Imagen, DALL-E 2, Latent Diffusion Models (LDM).

\subsection{Diffusion Models}
Diffusion models are models that iteratively restore $x_0$, the data from its Gaussian noise corruption $x_T$ in $T$ steps. This denoising step is called reverse diffusion process $p_\theta(x_{t-1} \mid x_t)$ and the reverse process is the forward diffusion which adds noise to the image written as  $q(x_t \mid x_{t-1})$.\newline
Both processes are modeled by the following equations: 

\begin{equation}
    q(x_t \mid x_{t-1}) = \mathcal{N}(x_t; \sqrt{\alpha_t} x_{t-1}, (1 - \alpha_t) \mathbf{I})
\end{equation}

\begin{equation}
   p_\theta(x_{t-1} \mid x_t) = \mathcal{N}(x_{t-1}; \mu_\theta(x_t, t), \sigma_\theta(x_t, t)) 
\end{equation}
The reverse diffusion process is modelled by a neural network and the loss function for this is calculated as the Mean Square Error (MSE) between two quantities, namely the true noise $\epsilon$ and the noise predicted by the neural network during reverse diffusion $\epsilon_\theta$. We denote the MSE by $L(\theta)$

\begin{equation}
  \mathcal{L}(\theta) = \mathbb{E}_{x_0, \epsilon, t} \left[ \left\| \epsilon - \epsilon_\theta(x_t, t) \right\|^2 \right]  
\end{equation}

\begin{figure}
    \centering
    \includegraphics[width=1\linewidth]{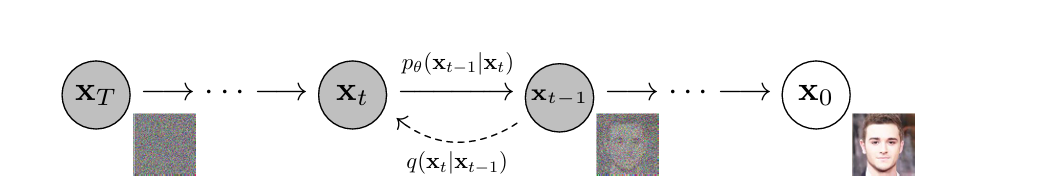}
    \caption{Diagramatic representation of the overall Diffusion Model's Process, originally shown in \cite{ho2020denoising}}
    \label{fig:enter-label}
\end{figure}

\subsection{Stable Diffusion Model}
Out of the various types of diffusion model, we have considered the stable diffusion model. Stable Diffusion is a large text-to-image diffusion model trained on billions of images. The reason we chose this is because of its ease of use and popularity.

Now there are three major components to a stable diffuion model which are as folllows:
\begin{itemize}
    \item A Variational Autoencoder (VAE)
    \item U-Net
    \item A text encoder - like CLIP
    \item Scheduler
\end{itemize}

\subsubsection{Variational Autoencoder (VAE)}
The VAE has two parts, namely the encoder and decoder. During this process, the encoder converts a large image into a low-dimensional representation of the image. This low-dimensional representation is called latent. 
Step-wise noise addition is done to these latents and these encoded latent representations of the images are passed as input to the U-Net model.
Convertion to latents reduces the memory, and the computational overhead is reduced.
The decoder transforms the latents back into image. In other words, the decoder converts the denoised latent to denoised image.

\subsubsection{U-Net}
 The U-Net is the Neural Network that predicts the denoised image representation of noisy latents. It takes in the noisy images, and predicts the noise added to the image and inturn the actual image can be got back by subtracting the noise from the noisy latent to get the denoised latent. 
It is known for the fact that it has two paths - a contractive and an expansive path. 
 The contractive path comprises of encoders that capture contextual information and reduce the resolution of the data. 
 The expansive path involves decoders that decode encoded data and use information from the contracting path via skip connections to generate a segmentation map.
The contracting path (which is the path of encoders), is responsible for identifying the relevant features in the input image.
While it is true that the encoders reduce the dimensionality of feature maps, they do increase the depth therefore capturing abstract representations of the input properly.
Decoding layers or Expansive Layers upsample the feature maps and the skip connections from the Contracting path (which is the Encoder path) helps to preserve spatial information lost in contracing path.

\subsubsection{Text Encoder - CLIP}

 The purpose of the text encoder is to transform the input prompt into embedding space which goes as input to the U-Net.
     The output of the Text Encoder acts as a guide for noisy latents when training the U-Net for the denoising process.
    The purpose of the encoder is to map a sequence of input tokens to a sequence of latent text-embeddings.
    Now stable diffusion uses an already pre-trained text encoder called CLIP, which stands for Contrastive Language-Image Pre-Training was introduced in 2021 in this paper \cite{clipPaper}.
    It is a joint image and text embedding model trained using around 400 million images and text pairs. This implies that it maps both the text and images to the same embedding space. This would mean the text prompt," An image of a dog" and an actual dog image would have similar embeddings and be close to each other in the vector space.

The CLIP model has two main components, a text encoder and an image encoder. The text encoder converts the text to vector embeddings and the image encoder converts the image to vector embeddings. For the image encoder, there are two variations commonly used - Vision Transformer (ViT) and ResNet-50. The text encoder is a transformer model. The model was trained on 400 million image, text pairs and the learning was done in a contrastive representation learning approach. In a contrastive representation learning approach, we aim to learn an embedding space in which similar sample pairs stay close to each other while the dissimilar ones are far apart. In a contrastive learning approach, we provide the model with input examples belonging to three types namely - anchor, positive, negative. These three types of inputs are converted to their vector embeddings. If we denote the distance between two vectors as $d$, then in this learning approach we ensure that $d(anchor,positive)$ is minimized and $d(anchor,negative)$ is maximized. Here in CLIP, the contrastive learning is done on the combination of image and text inputs. For example, the image of dog can be set as anchor, the positive can be set to the caption "Image of a dog", and negative to the caption "Image of a bird". So for N (image,text) pairs, the CLIP model is trained to predict which of the $N \times N$ pairings across a batch actually occurred by the joint training of the image and text encoder to maximise the cosine similarity of the text and image embeddings of the N actually similar pairs and minimize the cosine similairity of the $N^2-N$ incorrect pairings.

\begin{figure}
    \centering
    \includegraphics[width=0.8\linewidth]{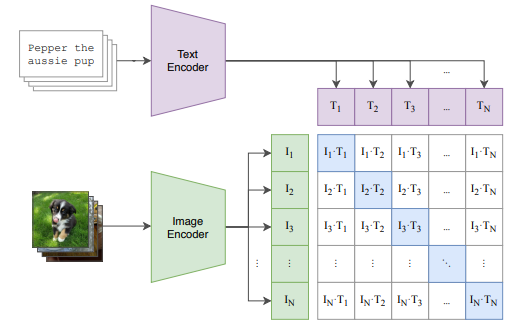}
    \caption{CLIP Model architecture as shown in \cite{clipPaper}}
    \label{fig:}
\end{figure}

\subsubsection{Scheduler}
A scheduler controls how the noise is to be removed step-by-step during the de-noisning step i.e. the reverse diffusion process. Given the number of steps for the reverse diffusion process to happen, it determines how much noise is to be removed at each step.

\subsection{Machine Unlearning in Diffusion Models}

The task of unlearning in diffusion models has recently gained prominence due to growing concerns around privacy, fairness, intellectual property, and the generation of NSFW (Not Safe For Work) content. Several recent works have proposed different strategies to address the challenge of selectively removing or modifying specific concepts within trained diffusion models, particularly stable diffusion.

Forget-Me-Not \cite{zhang2023forgetmenot}  introduces a novel method to forget specific concepts in diffusion models by suppressing the cross-attention scores between token embeddings and corresponding visual features. This attention-level manipulation enables the model to erase target concepts while retaining unrelated ones. To evaluate the effectiveness of forgetting, the authors propose a metric called the Memorization Score (M-Score), which measures the change in cosine similarity of concept embeddings before and after unlearning. Furthermore, the use of Textual Inversion enhances generality, especially in cases where the target concepts are not easily describable using standard text prompts. Their experiments demonstrate effectiveness in multi-concept forgetting and concept correction, allowing previously overshadowed prompts to emerge during generation.

In a conceptually distinct approach, \cite{wu2024scissorhands} employs a two-phase process involving a Trimming Phase and a Repairing Phase. During the trimming phase, connection sensitivity is used to identify and reinitialize the top k\% of model parameters most influenced by the forget data. Subsequently, the repairing phase adopts a min-max optimization strategy, maximizing the loss on forget data while minimizing it on retain data. This enables fast and effective unlearning, especially when partial access to the original training data is available. 

The EraseDiff framework proposed in \cite{erasediff}presents a constrained optimization formulation that rewrites the bi-level optimization problem into a single objective. Inspired by \cite{wu2024scissorhands} Scissorhands, this approach updates model parameters using a gradient-based strategy where the losses from forget and retain data are simultaneously minimized in opposite directions. 

A critical bottleneck in unlearning is the availability of the forget dataset itself. Addressing this, UnlearnCanvas \cite{zhang2024unlearncanvas} introduces a benchmark dataset designed to evaluate unlearning methods on image generation models by focusing on stylistic forget tasks. The authors evaluate nine unlearning methods against four criteria—unlearning accuracy, retainability, FID score, and efficiency. Their results highlight the absence of a single best-performing method across all metrics, thereby motivating further research on balanced unlearning.

Another notable direction is targeted unlearning at the text encoder level, as proposed in \cite{Fuchi_2024_BMVC}. Instead of modifying the entire diffusion pipeline, the authors focus on the CLIP text encoder to erase the influence of specific textual concepts. This few-shot unlearning approach is shown to maintain the quality of generated images better, as it avoids disrupting the broader visual decoding architecture. Using a small number of real images, the method effectively reduces concept influence while preserving general model performance.

Collectively, these studies underscore the complexity of selective unlearning in diffusion models and highlight trade-offs between effectiveness, computational efficiency, and retention of unrelated knowledge. The diversity in methodologies—from attention suppression to parameter reinitialization and constrained optimization—reflects the multifaceted nature of the problem and paves the way for future innovations in responsible generative modeling.

\section{Proposed Method}
\subsection{Unlearning in Diffusion Model Pipeline}

Concept Unlearning in image generation models is crucial due to the need for forgetting data from image generation models without the need to retrain, as retraining the diffusion models is extremely computationally intensive. 
\subsubsection{Prompt Conditioning with CLIP}

Text prompts are embedded into feature vectors using the CLIP text encoder. These vectors are then used to guide the reverse diffusion process, ensuring that the generated image aligns with the semantic content of the prompt. The CLIP model contains a projection matrix 
\( P \in \mathbb{R}^{d \times d} \) that maps intermediate token features to final embeddings.

\subsubsection{Low-Rank Adaptation for Concept Forgetting}

Concept forgetting is achieved by perturbing the projection matrix \( P \) using a low-rank matrix \( \Delta P = AB^\top \), where \( A, B \in \mathbb{R}^{d \times r} \) and \( r \ll d \). This ensures changes are efficient and restricted to a low-dimensional subspace. This allows for Parameter Efficiency, reducing the number of learnable parameters, targeted unlearning along with stability for the retain set.

For selective forgetting of concepts within the Stable Diffusion pipeline, we propose a low-rank perturbation method applied to the projection matrix of the CLIP text encoder. This method constrains modifications to a low-dimensional subspace, thereby ensuring stability and efficiency while minimizing disruption to retained concepts.


\begin{figure*}[t]
\centering
\includegraphics[width=0.8\textwidth]{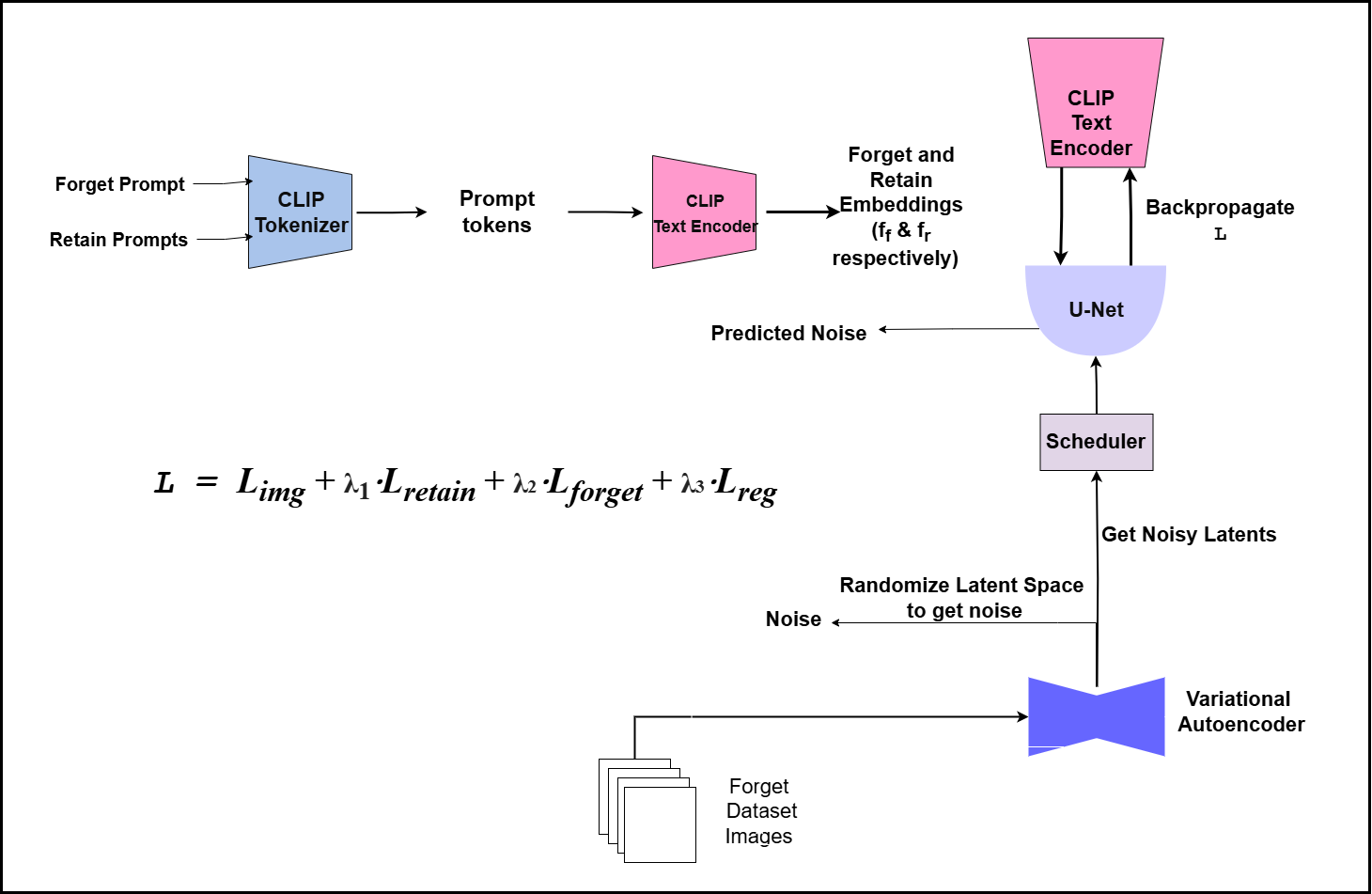}
\caption{Unlearning Process of the Stable Diffusion Pipeline}
\label{fig:diffusionunlearnalgo}
\end{figure*}


The procedure begins with the initialization of two low-rank matrices \( A, B \in \mathbb{R}^{d \times r} \), such that their product \( \Delta P = AB^\top \) defines a learnable perturbation of the original projection matrix \( P \in \mathbb{R}^{d \times d} \). The rank \( r \) is chosen to be small (typically between 4 and 8), allowing for expressive yet compact updates.

The VAE, U-Net, and most of the CLIP-based text encoder components are frozen to preserve the integrity of the pretrained model. Only the final normalization layer and the last few encoder layers of the text encoder remain trainable. This allows us to only change the last few layers of the text encoder model during backpropagation in order to avoid disturbing the retain space.

During each training iteration, images from the forget dataset \( \mathcal{D}_f \) are encoded into latent representations using the pretrained VAE encoder. A noise level \( t \) is sampled uniformly to simulate the forward diffusion process, and noise is added to the latent using a diffusion scheduler. Corresponding text prompts for the concept to be forgotten (\( x_f \)) and the prompts to be retained (\( x_r \)) are embedded using the CLIP text encoder,yielding feature representations \( f_f \) and \( f_r \), respectively.

The noised latents and the forget concept embeddings \( f_f \) are input to the U-Net, producing predicted noise estimates \( \hat{z} \). The reconstruction fidelity is enforced by a standard noise prediction loss, such as the mean squared error (MSE) between \( \hat{z} \) and the originally added noise.

\begin{algorithm}[H]
\caption{Low-Rank Concept Unlearning in Stable Diffusion}
\label{algo:unlearn}
\begin{algorithmic}[1]
\Require Text encoder $C$, projection matrix $P \in \mathbb{R}^{d \times d}$, prompts $x_f$, $x_r$, forget dataset $\mathcal{D}_f$, VAE, UNet, rank $r$
\State Initialize low-rank matrices $A, B \in \mathbb{R}^{d \times r}$ randomly
\State Keep rank $r$ as a single-digit multiple of 2 (preferably 4--8)
\State Freeze VAE, UNet, and most of $C$; unfreeze last encoder layers and final norm layer of CLIP text model
\State $F_{\text{forget}} \sim \mathcal{N}(0, 2.0)$, \quad $n_{\text{epochs}} \gets 10$
\For{each epoch}
  \For{each image $I \in \mathcal{D}_f$}
    \State $z \gets \text{VAE.encode}(I)$
    \State $t \sim \mathcal{U}[0, T]$
    \State $z_{\text{noisy}} \gets \text{Scheduler.add\_noise}(z, t)$
    \State $f_f \gets C(x_f)$, \quad $f_r \gets C(x_r)$
    \State $\hat{z} \gets \text{UNet}(z_{\text{noisy}}, t, f_f)$
    \State $\mathcal{L}_{\text{img}} \gets -\text{MSE}(\hat{z}, \text{noise})$
    \State $\Delta P \gets AB^\top$
    \State $\mathcal{L}_{\text{retain}} \gets \text{MSE}((P + \Delta P) f_r^\top, Pf_r^\top)$
    \State $\mathcal{L}_{\text{forget}} \gets \text{MSE}((P + \Delta P) f_f^\top, F_{\text{forget}})$
    \State $\mathcal{L}_{\text{reg}} \gets \|\Delta P\|_F$
    \State $\mathcal{L}_{\text{total}} \gets \mathcal{L}_{\text{img}} + \lambda_1 \mathcal{L}_{\text{retain}} + \lambda_2 \mathcal{L}_{\text{forget}} + \lambda_3 \mathcal{L}_{\text{reg}}$
    \State Backpropagate $\mathcal{L}_{\text{total}}$ and update $A, B$, and trainable layers of $C$
  \EndFor
\EndFor
\State \Return Modified text encoder $C$, low-rank components $A, B$

\end{algorithmic}
\end{algorithm}

Let the projection matrix be \( P \in \mathbb{R}^{d \times d} \), the perturbation \( \Delta P \in \mathbb{R}^{d \times d} \), 
retained features \( f_r \in \mathbb{R}^{n_r \times d} \), forgotten features \( f_f \in \mathbb{R}^{n_f \times d} \), 
and the target forget representation be \( F_{\text{forget}} \in \mathbb{R}^{d \times n_f} \).  
The loss components are defined as follows:


\begin{align*}
\mathcal{L}_{\text{img}} &= -\text{MSE}(\hat{z}, \text{noise}) \\
\mathcal{L}_{\text{retain}} &= \text{MSE}((P + \Delta P) f_r^\top, P f_r^\top) \\
\mathcal{L}_{\text{forget}} &= \text{MSE}((P + \Delta P) f_f^\top, F_{\text{forget}})
\end{align*}

Let \( \mathcal{L}_{\text{img}} \) be the image reconstruction or generation loss.  
Then, the total loss is given by:
\vspace{-0.01em}
\begin{equation}
\mathcal{L} = \mathcal{L}_{\text{img}} + \lambda_1 \cdot \mathcal{L}_{\text{retain}} + \lambda_2 \cdot \mathcal{L}_{\text{forget}} + \lambda_3 \cdot \mathcal{L}_{\text{reg}}
\end{equation}
\vspace{-0.01em}
Training proceeds for a small number of epochs (typically ten), and updates are applied only to the low-rank matrices \( A, B \) and the unfrozen layers of the text encoder. The final output of the procedure is a modified text encoder capable of forgetting the undesired concept while preserving general generation capabilities and the semantics of retained prompts.
\vspace{-0.1em}\newline
Algorithm\ref{algo:unlearn} gives us a parameter-efficient way to forget concepts for a diffusion model pipeline without disrupting the retain prompts. By using low-rank projections to perturb the image space, we achieve targeted unlearning while retaining overall generative quality. 

\vspace{-0.1em}

\section{Experiments}
All experiments were conducted using the Google Colab TPU environment to leverage accelerated hardware for model fine-tuning and unlearning procedures. We employed three prominent unlearning techniques applied to Stable Diffusion models: \cite{erasediff}EraseDiff, \cite{Fuchi_2024_BMVC}Few-Shot Unlearning, and \cite{gandikota2024unified}Unified Concept Editing (UCE). These approaches were used to selectively remove specific concepts from the model’s generative capabilities.

To assess the effectiveness of unlearning, we targeted three distinct concepts for removal: Taj Mahal, Mickey Mouse, and Siberian Husky. Each unlearning method was evaluated independently on these concepts. A rank of 8 was selected for all experiments. This rank was chosen as a balanced trade-off: it is sufficiently low to ensure computational efficiency and preserve unrelated (retain) concepts, while still being high enough to enable effective forgetting of the target (forget) concepts.

\subsection{Metrics}
This section explains the different Metrics used to evaluate the unlearning in the diffusion model with respect to the different concepts unlearned. Table \ref{Unlearning-Results} shows the overview of the results we got for these metrics across all concepts.

\subsubsection{CLIP Scores}
CLIP score is used to measure the  similarity between the image and textual description by calculating the cosine similarity between the image and text embeddings. 
A higher CLIP score indicates a higher similarity between the image and text embeddings.
Retain CLIP measures how well the model retains knowledge it was not supposed to forget during unlearning and the image and textual description are well aligned for the retain concept.
Forget CLIP score measures how well the model forgets the target concept. A lower forget CLIP score means the model has successfully unlearned or removed that concept.

\subsubsection{FID Score}
Fréchet Inception Distance (FID) is a metric used to evaluate the quality of generated images in generative models like GANs, diffusion models or in any other image generative models.
FID compares the distribution of real images with the distribution of generated images in the feature space i.e. the vector space.     
This is done by taking both, the real set of images and the generated set of images through a pre-trained Inception v3 network. The activations are taken from deep layers and they are treated as multi-variate gaussians and the distance between the two gaussians of the real and generated image is computed. Lower FID means the generated images are closer to the real images (ground truth images). 
Since , we are unlearning a concept, we want the generated image to be far away as possible from the ground truth and hence for us, a higher FID is desirable.

\subsubsection{Detection Rate}
Detection Rate measures the model's ability to correctly identify all relevant instances of a particular class.
It is defined as the ratio of the True Positives to the sum of True Positives and False Negatives. For an unlearning task, a good detection rate is to be as close to zero as possible.

\subsubsection{Unlearning Time}
It is the overall time taken for the unlearning algorithm to run in order to  remove a specific concept from the model.

\section{Results}

\begin{figure}[H]
    \centering
    \begin{subfigure}[t]{0.48\textwidth}
        \centering
        \includegraphics[width=0.5\linewidth]{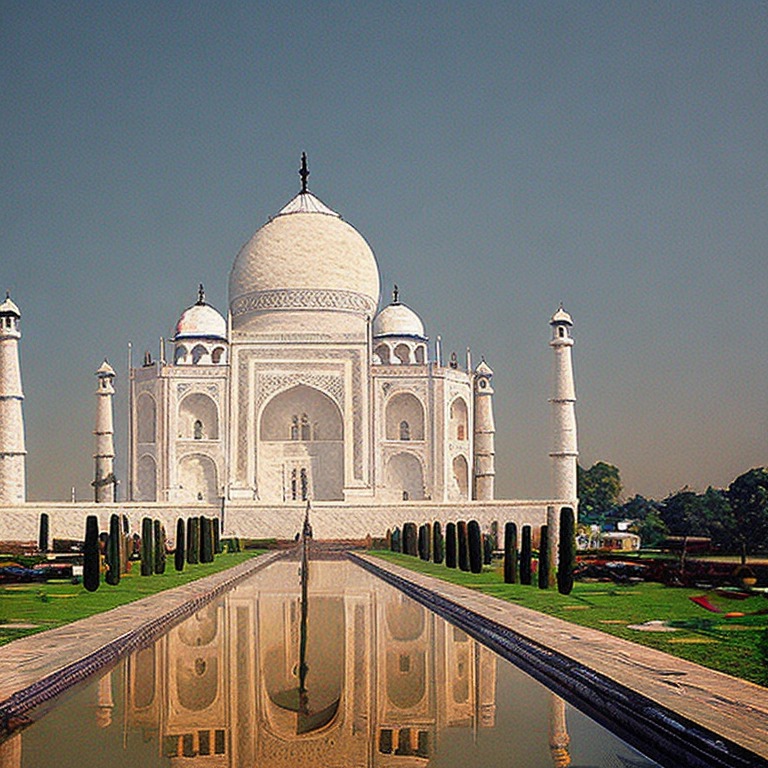}
        \caption{Before unlearning}
        \label{fig:tajbefore}
    \end{subfigure}
    \hfill
    \begin{subfigure}[t]{0.48\textwidth}
        \centering
        \includegraphics[width=0.5\linewidth]{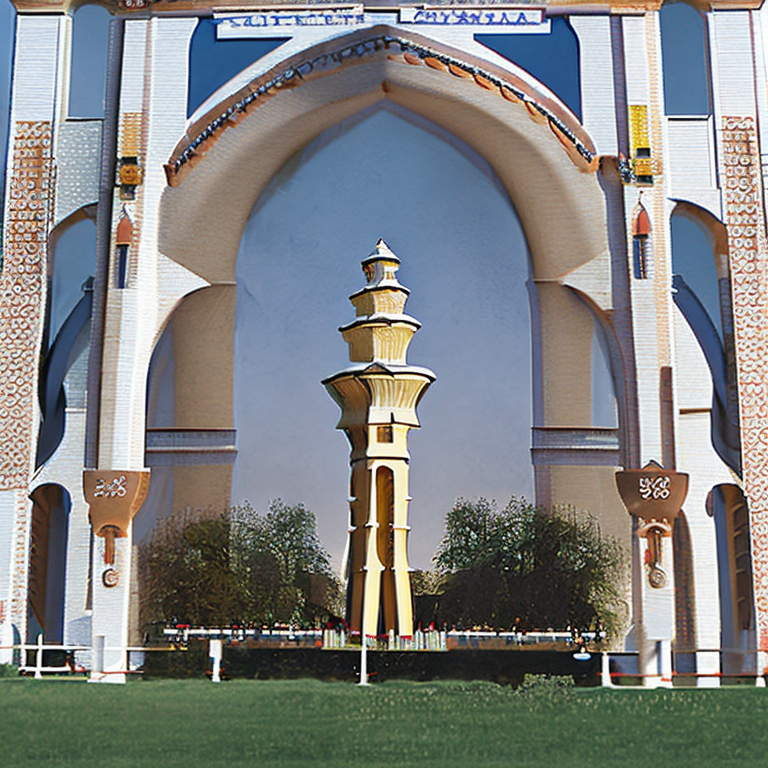}
        \caption{After unlearning}
        \label{fig:taj_after}
    \end{subfigure}
    \caption{Generated images for the prompt: "An image of the Taj Mahal" before and after unlearning.}
    \label{fig:Tajunlearn}
\end{figure}

\begin{figure}[H]
    \centering
    \begin{subfigure}[t]{0.48\textwidth}
        \centering
        \includegraphics[width=0.5\linewidth]{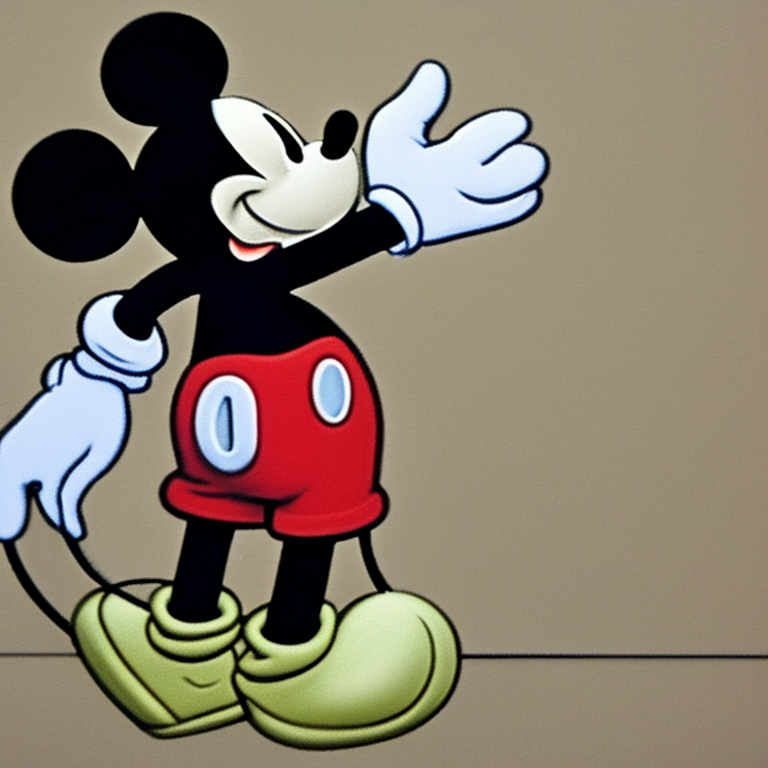}
        \caption{Before unlearning}
        \label{fig:mickey_before}
    \end{subfigure}
    \hfill
    \begin{subfigure}[t]{0.48\textwidth}
        \centering
        \includegraphics[width=0.5\linewidth]{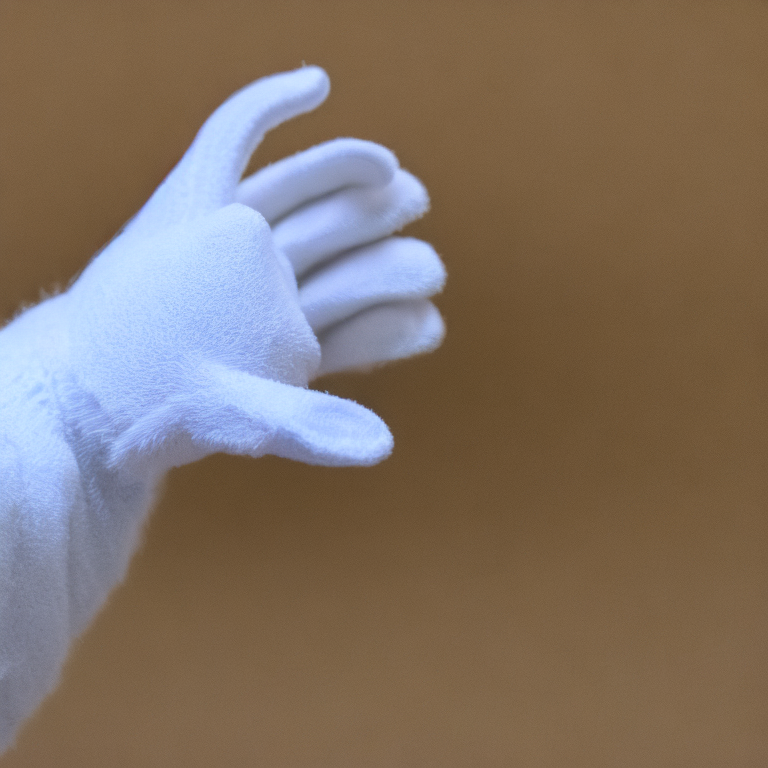}
        \caption{After unlearning}
        \label{fig:mickey_after}
    \end{subfigure}
    \caption{Generated images for the prompt: "Mickey Mouse with white gloves" before and after unlearning.}
    \label{fig:mickeyunlearn}
\end{figure}

\begin{figure}[H]
    \centering
    \begin{subfigure}[t]{0.48\textwidth}
        \centering
        \includegraphics[width=0.5\linewidth]{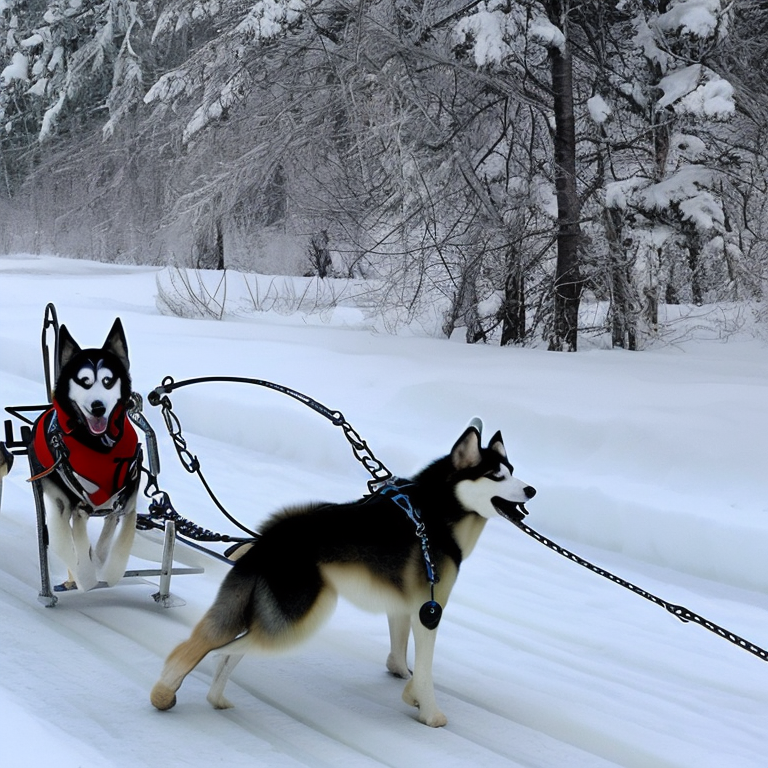}
        \caption{Before unlearning}
        \label{fig:mickey_before}
    \end{subfigure}
    \hfill
    \begin{subfigure}[t]{0.48\textwidth}
        \centering
        \includegraphics[width=0.5\linewidth]{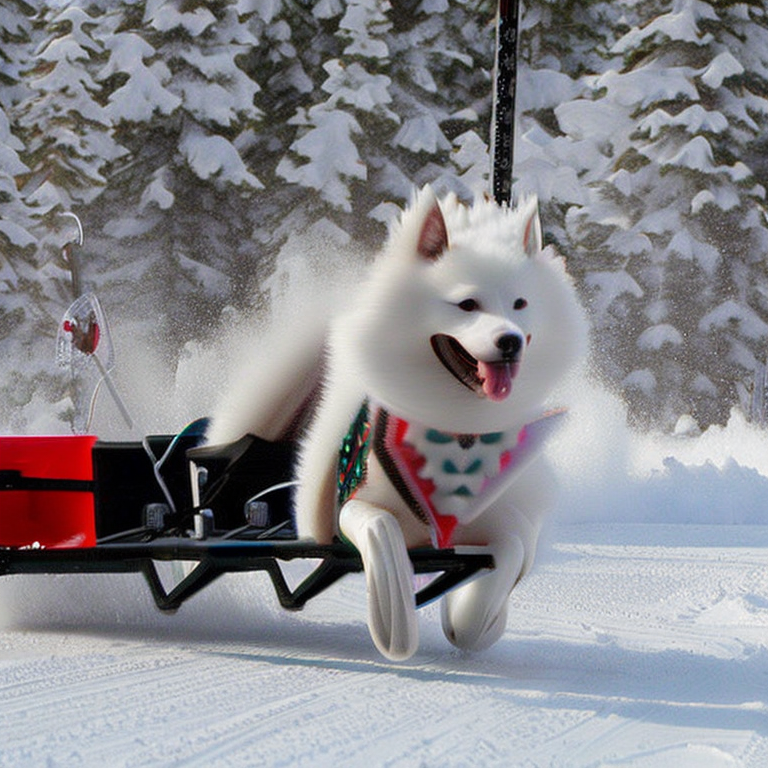}
        \caption{After unlearning}
        \label{fig:mickey_after}
    \end{subfigure}
    \caption{Generated images for the prompt: "A Siberian Husky pulling the sled during winter" before and after unlearning.}
    \label{fig:Huskyunlearn}
\end{figure}

\subsection{Loss Graphs}
Loss vs epoch graphs were drawn for unlearning each of the three concepts. Three types of losses were calculated, which is the retain loss, forget loss, and average loss. Average loss is the average of the total loss across each epoch. Retain Loss is calculated with respect to the Retain concept and Forget Loss is calculated with respect to the Forget concept. The ideal retain loss is 0 and forget loss should be as high as possible, and for each epoch the retain loss should be lower than the forget loss. In all three concepts, we have obtained ideal retain losses.






\begin{figure}[H]
    \centering

    \begin{subfigure}[b]{1\linewidth}
        \centering
        \includegraphics[width=\linewidth]{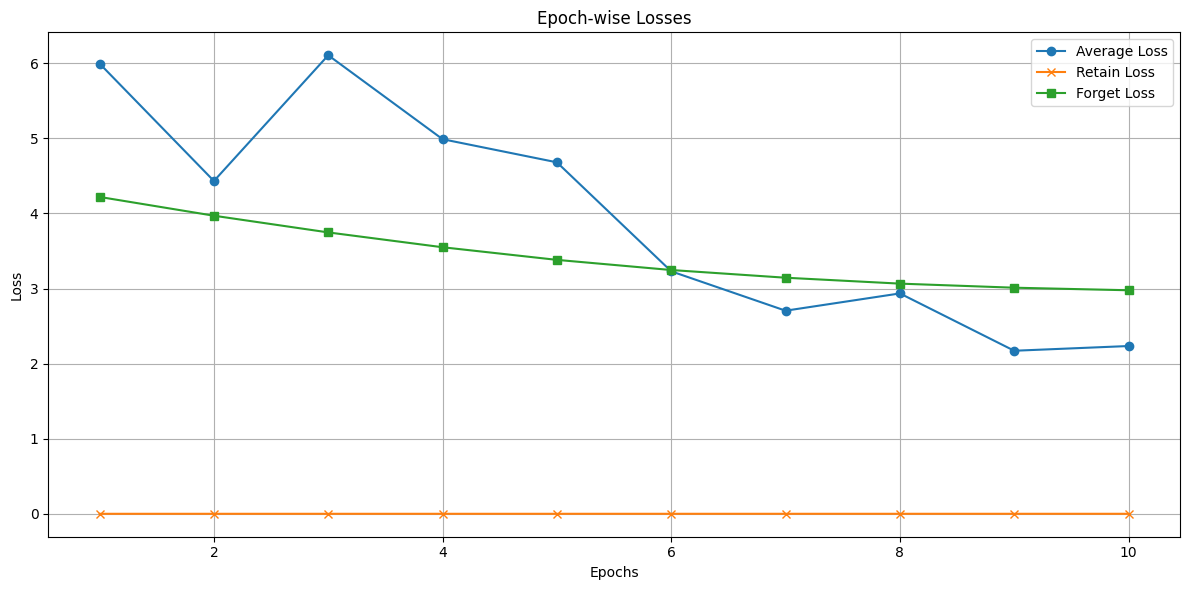}
        \caption{Taj Mahal}
        \label{fig:loss-taj-mahal}
    \end{subfigure}
    
    \vspace{0.5cm}

    \begin{subfigure}[b]{1\linewidth}
        \centering
        \includegraphics[width=\linewidth]{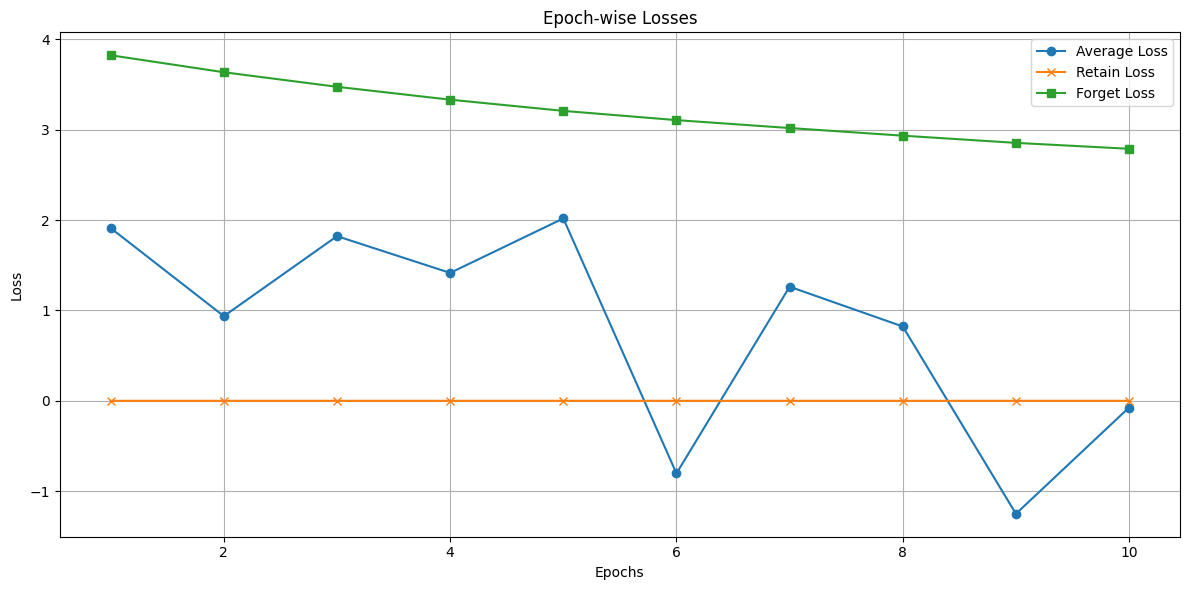}
        \caption{Mickey Mouse}
        \label{fig:loss-mickey-mouse}
    \end{subfigure}
    
    \vspace{0.5cm}

    \begin{subfigure}[b]{1\linewidth}
        \centering
        \includegraphics[width=\linewidth]{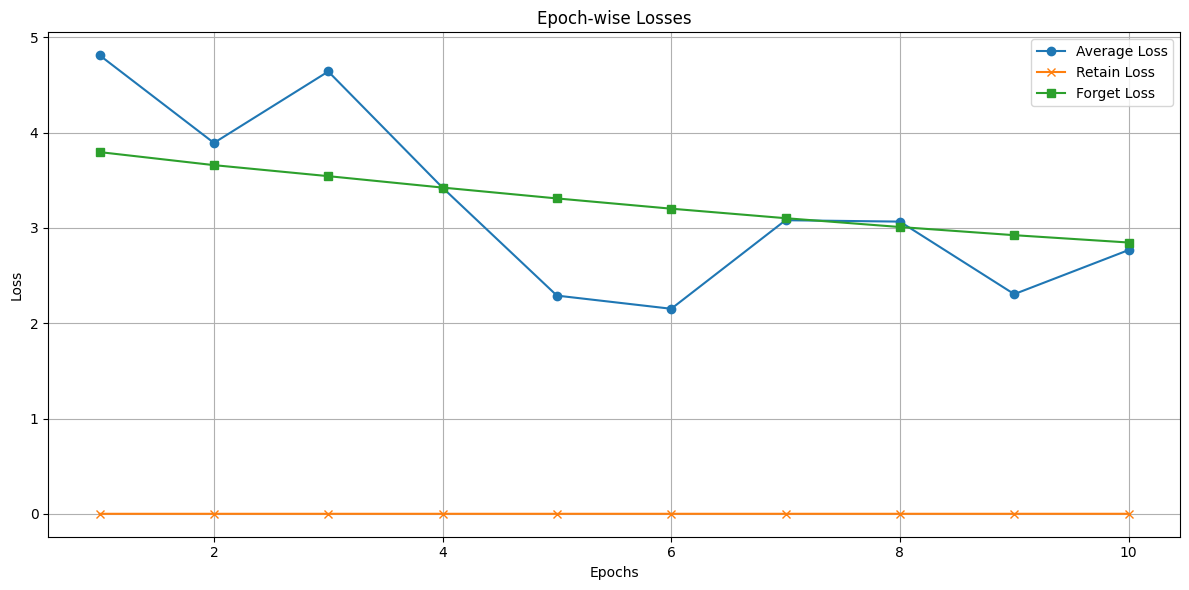}
        \caption{Siberian Husky}
        \label{fig:loss-siberian-husky}
    \end{subfigure}

    \caption{Loss Graphs for unlearning different concepts showing the epoch-wise losses}
    \label{fig:combined-loss-graphs}
\end{figure}


\begin{table}[H]
\centering
\resizebox{\columnwidth}{!}{%
\begin{tabular}{|l|c|c|c|}
\hline
\textbf{Metric} & \textbf{Siberian Husky} & \textbf{Taj Mahal} & \textbf{Mickey Mouse} \\
\hline
Retain CLIP & 0.3294 & 0.3161 & 0.3478 \\
Forget CLIP & 0.2938 & 0.2590 & 0.2934 \\
Unlearning Time (s) & 52.5 & 51.1 & 58.2 \\
FID Score & 151.611 & 343.002 & 342.287 \\
Detection Rate & 0.01 & 0 & 0.02 \\
\hline
\end{tabular}%
}
\caption{Unlearning Results for Different Concepts}
\label{Unlearning-Results}
\end{table}

Going by the above metrics, it is observed that our proposed unlearning method runs in very less time of less than one minute which is good. Moreover seeing the earlier figures, we can observe that the concept is truly unlearned and influence of it is removed, such that when generating with secondary objects only the secondary objects are generated e.g. the prompt "Mickey Mouse with white gloves" generates an image consisting only of the white gloves.

\ref{Unlearning:Baselines} are the results obtained by running baselines with the metrics averaged across the three concepts we have run to forget.

\begin{table}[H]
\begin{tabular}{|l|l|l|l|}
\hline
                            & UCE \cite{gandikota2024unified} & EraseDiff \cite{erasediff} & Few-Shot \cite{Fuchi_2024_BMVC} \\ \hline
Forget CLIP        & 0.346 & 0.319                & 0.295            \\ \hline
Unlearning Time (s)& 630.5& 786.45            & 76.2             \\ \hline
FID Score           & 57.35 & 1.42              & 174.32           \\ \hline
Detection Rate      & 0.13  & 0.05              & 0.03             \\ \hline
\end{tabular}
\caption{Baselines implemented}
\label{Unlearning:Baselines}
\end{table}

\section{Conclusion}
In this paper we discuss a novel pipeline for unlearning in diffusion models by looking into Stable Diffusion by unlearning concepts ranging from famous characters to landmark monuments through the CLIP encoder, which is a part of the Stable Diffusion pipeline. The unlearning process is very small with times of around one minute. We have got minimum CLIP scores for Forget set and maximum values for Retain. Furthermore, our FID scores are much high indicating a very minute co-relation with the ground truth images of concepts thus showing succesful unlearning. The fact that our unlearning approach takes around 1 minute is also a good indicator that our unlearning approach is efficient.
Future works that can be done in this direction can involve adding more few shot images and observing how the scores of these metrics change. Furthermore, testing on mixed images consisting of both forget and retain concepts would be a good way of further understanding generality of the unlearned model.

A major breakthrough in future would be to further develop this framework to become truly zero-shot in the same unlearning time amount, thereby reducing the need of forget data from few to none. While our method can be optimized to be zero shot by simply randomizing the latents instead of encoding through the Auto-Encoder, it requires much more time to run to truly get any substantial results. 


\bibliographystyle{IEEEtran}
\bibliography{references}

\begin{thebibliography}{10}
\providecommand{\url}[1]{#1}
\csname url@samestyle\endcsname
\providecommand{\newblock}{\relax}
\providecommand{\bibinfo}[2]{#2}
\providecommand{\BIBentrySTDinterwordspacing}{\spaceskip=0pt\relax}
\providecommand{\BIBentryALTinterwordstretchfactor}{4}
\providecommand{\BIBentryALTinterwordspacing}{\spaceskip=\fontdimen2\font plus
\BIBentryALTinterwordstretchfactor\fontdimen3\font minus \fontdimen4\font\relax}
\providecommand{\BIBforeignlanguage}[2]{{%
\expandafter\ifx\csname l@#1\endcsname\relax
\typeout{** WARNING: IEEEtran.bst: No hyphenation pattern has been}%
\typeout{** loaded for the language `#1'. Using the pattern for}%
\typeout{** the default language instead.}%
\else
\language=\csname l@#1\endcsname
\fi
#2}}
\providecommand{\BIBdecl}{\relax}
\BIBdecl

\bibitem{voigt2017eu}
P.~Voigt and A.~Von~dem Bussche, ``The eu general data protection regulation (gdpr),'' \emph{A Practical Guide, 1st Ed., Cham: Springer International Publishing}, vol.~10, no. 3152676, pp. 10--5555, 2017.

\bibitem{goldman2020introduction}
E.~Goldman, ``An introduction to the california consumer privacy act (ccpa),'' \emph{Santa Clara Univ. Legal Studies Research Paper}, 2020.

\bibitem{lexica2025}
{Lexica}, ``Lexica - ai art generator and search tool,'' \url{https://lexica.art/}, 2025, accessed: 2025-05-03.

\bibitem{novelai2025}
{NovelAI}, ``Novelai - ai-assisted storytelling and art generation,'' \url{https://novelai.net/}, 2025, accessed: 2025-05-03.

\bibitem{picsart2025}
{Picsart Inc.}, ``Picsart - ai photo and video editing tools,'' \url{https://picsart.com/}, 2025, accessed: 2025-05-03.

\bibitem{firefly2025}
{Adobe Inc.}, ``Adobe firefly - ai generative creative tools,'' \url{https://www.adobe.com/products/firefly.html}, 2025, accessed: 2025-05-03.

\bibitem{gandikota2023erasing}
R.~Gandikota, J.~Materzynska, J.~Fiotto-Kaufman, and D.~Bau, ``Erasing concepts from diffusion models,'' in \emph{Proceedings of the IEEE/CVF International Conference on Computer Vision}, 2023, pp. 2426--2436.

\bibitem{Fuchi_2024_BMVC}
\BIBentryALTinterwordspacing
M.~Fuchi and T.~Takagi, ``Erasing concepts from text-to-image diffusion models with few-shot unlearning,'' in \emph{35th British Machine Vision Conference 2024, {BMVC} 2024, Glasgow, UK, November 25-28, 2024}.\hskip 1em plus 0.5em minus 0.4em\relax BMVA, 2024. [Online]. Available: \url{https://papers.bmvc2024.org/0216.pdf}
\BIBentrySTDinterwordspacing

\bibitem{gal2022image}
R.~Gal, Y.~Alaluf, Y.~Atzmon, O.~Patashnik, A.~H. Bermano, G.~Chechik, and D.~Cohen-Or, ``An image is worth one word: Personalizing text-to-image generation using textual inversion,'' \emph{arXiv preprint arXiv:2208.01618}, 2022.

\bibitem{ho2020denoising}
J.~Ho, A.~Jain, and P.~Abbeel, ``Denoising diffusion probabilistic models,'' \emph{Advances in neural information processing systems}, vol.~33, pp. 6840--6851, 2020.

\bibitem{clipPaper}
A.~Radford, J.~Kim, C.~Hallacy, A.~Ramesh, G.~Goh, S.~Agarwal, G.~Sastry, A.~Askell, P.~Mishkin, J.~Clark, G.~Krueger, and I.~Sutskever, ``Learning transferable visual models from natural language supervision,'' 02 2021.

\bibitem{zhang2023forgetmenot}
E.~Zhang, K.~Wang, X.~Xu, Z.~Wang, and H.~Shi, ``Forget-me-not: Learning to forget in text-to-image diffusion models,'' \emph{arXiv preprint arXiv:2211.08332}, 2023.

\bibitem{wu2024scissorhands}
J.~Wu and M.~Harandi, ``Scissorhands: Scrub data influence via connection sensitivity in networks,'' \emph{arXiv preprint arXiv:2401.06187}, 2024.

\bibitem{erasediff}
J.~Wu, T.~Le, M.~Hayat, and M.~Harandi, ``Erasediff: Erasing data influence in diffusion models,'' \emph{arXiv preprint arXiv:2401.05779}, 2024.

\bibitem{zhang2024unlearncanvas}
Y.~Zhang, C.~Fan, Y.~Zhang, Y.~Yao, J.~Jia, J.~Liu, G.~Zhang, G.~Liu, R.~R. Kompella, X.~Liu \emph{et~al.}, ``Unlearncanvas: Stylized image dataset for enhanced machine unlearning evaluation in diffusion models,'' \emph{arXiv preprint arXiv:2402.11846}, 2024.

\bibitem{gandikota2024unified}
R.~Gandikota, H.~Orgad, Y.~Belinkov, J.~Materzy{\'n}ska, and D.~Bau, ``Unified concept editing in diffusion models,'' in \emph{Proceedings of the IEEE/CVF Winter Conference on Applications of Computer Vision}, 2024, pp. 5111--5120.

\end{thebibliography}

\end{document}